\documentclass[entropy,article,accept,moreauthors,pdftex]{mdpi}

\firstpage{1}
\pubvolume{21}
\issuenum{8}
\articlenumber{723}
\pubyear{2019}
\copyrightyear{2019}
\history{Received: 4 July 2019; Accepted: 24 July 2019; Published: 25 July 2019}
\updates{yea}
\usepackage{subcaption}

\Title{Coin.AI: A Proof-of-Useful-Work Scheme for Blockchain-Based Distributed Deep Learning}

\Author{Alejandro Baldominos *\orcidA{} and Yago Saez \orcidB{}}

\AuthorNames{Alejandro Baldominos and Yago Saez}

\address[1]{Computer Science Department, Universidad Carlos III of Madrid, 28911 Leganés, Madrid, Spain}

\corres{Correspondence: abaldomi@inf.uc3m.es}

\abstract{One decade ago, Bitcoin was introduced, becoming the first cryptocurrency and establishing the concept of ``blockchain'' as a distributed ledger. As of today, there are many different implementations of cryptocurrencies working over a blockchain, with different approaches and philosophies. However, many of them share one common feature: they require proof-of-work to support the generation of blocks (mining) and, eventually, the generation of money. This proof-of-work scheme often consists in the resolution of a cryptography problem, most commonly breaking a hash value, which can only be achieved through brute-force. The main drawback of proof-of-work is that it requires ridiculously large amounts of energy which do not have any useful outcome beyond supporting the currency. In this paper, we present a theoretical proposal that introduces a proof-of-useful-work scheme to support a cryptocurrency running over a blockchain, which we named Coin.AI. In this system, the mining scheme requires training deep learning models, and a block is only mined when the performance of such model exceeds a threshold. The distributed system allows for nodes to verify the models delivered by miners in an easy way (certainly much more efficiently than the mining process itself), determining when a block is to be generated. Additionally, this paper presents a proof-of-storage scheme for rewarding users that provide storage for the deep learning models, as well as a theoretical dissertation on how the mechanics of the system could be articulated with the ultimate goal of democratizing access to artificial intelligence.}


\begin{document}

%
%
\section{Introduction}
In 2008, a researcher (or maybe a group of researchers) hidden under the pseudonym ``Satoshi Nakamoto'' published a paper describing a peer-to-peer digital money system \citep{Nakamoto08Bitcoin}. One year later, this paper would give birth to the Bitcoin network. As of today, Bitcoin is the most used cryptocurrency, although many other alternatives have arisen that are based on similar principles \citep{Mukhopadhyay16_Cryptos,Tschorsch16_Cryptos}. Notwithstanding, the volume of daily operations is of about 290,000 \citep{BitcoinCharts}, an amount which is still very far away of the number of VISA operations, which would exceed the 280 million daily transactions \citep{Velde13,McCallum15}.

In spite of the relatively small number of transactions, Bitcoin is a cryptocurrency listed in the currency market, and can be exchanged for dollars, euros or other fiat currency. Its value in dollars has increased significantly since its appearance in 2009, turning into one of the products with highest financial performance for investors. As an example, by mid-2010 the value of the bitcoin was established in about US\$0.06, whereas as of July 2019 it is valued around US\$12,000, an increase of 200,000 times its value in a period of 10 years.

Additionally, in recent years many other cryptocurrencies have appeared whose workings are very similar to Bitcoin, although with some particularities and different philosophies. Probably, the most significant example is the Ether, the currency provided by the Ethereum platform, which was presented in 2013 by \citet{Buterin13}. Again, this currency has seen its value increased significantly since it was born in July 2015, traded at about US\$0.70 by August 2015 and with a value of more than US\$220 by July 2019.

Bitcoin, Ethereum, and many other cryptocurrencies of this kind work atop a technology commonly known as \textit{blockchain} \cite{Iansiti17}. Blockchain technology allows to store transactions in blocks, which are placed sequentially, thus forming a chain. A cryptographic mechanism prevents the chain from being tampered, and the only operation that can alter the blockchain is appending a new block at the end of it. In other words, this mechanism avoids malicious activities that can try to alter or remove transactions that took place at some time in the past. Blockchains have been widely used with academic purposes, such as developing consensus protocols \cite{Garay17} or developing data exchange schemas for the internet of things \cite{Saia19}. Noticeably, they have also been used for not so honest purposes, and particularly some financial frauds, such as Ponzi schemes, have been detected in public networks \cite{Bartoletti17}.

Although a blockchain can be private and centralized, most blockchains standing behind a cryptocurrency such as Bitcoin or Ether are often public and decentralized. In general, a public blockchain allows everyone to query the list of transactions, providing full transparency about the contents of the ledger. Additionally, by being decentralized, a blockchain stores the blocks and validates and confirms transactions using a distributed network of computer systems.

The fact of having a decentralized system is key in these types of cryptocurrencies blockchains. By not relying on a centralized system that runs the platform code, there must be an incentive for users to join the network that is supporting the blockchain, and without whom transactions could not be done, putting an end to the cryptocurrency.

Nowadays, this incentive is the block mining. In summary, a block can only be appended to the blockchain if it has been mined by a user. The mining process often consists in solving a cryptographic problem whose resolution is computationally expensive and requires brute-forcing, but for which given a solution, it is easy to validate. This cryptographic problem often consists in finding a number, called \textit{nonce}, which must be appended to the transactions that are about to be included in the block to be mined. Next, the value resulting from applying a hash function over the concatenation of the hash of the previous block, the list of transactions and the nonce will be computed. The purpose will be to find a \textit{nonce} so that such hash value will be smaller than a given hash. Because hash functions are computed only in one direction (i.e., it is impossible to compute the original value from the hash value), there are no formal procedures for solving this problem, other than brute-force, i.e., trying different nonce values sequentially until a hash fulfilling the condition is found.

Once a user has successfully mined a block, he obtains a reward, which is provided by the network. As an example, mining a block in the Bitcoin network is currently rewarded with a total of 12.5 bitcoins (this value is estimated to halve by May 2020). In addition, the user mining the block is rewarded with the fees involved in the transactions that are included in the block. In most current cryptocurrencies schemes, these fees are not compulsory but highly recommended for a transaction to be included faster in the blockchain: this is because miners are responsible for choosing the transactions included in a block, and it is likely that they will choose those with larger fees.

This block mining scheme receives the name of \textit{proof-of-work}, since it requires to have performed a computational work (in this case, brute-force for breaking the cryptographic problem) in order to be able to generate a new block to be appended to the blockchain and thus receive a reward. The dynamics of mining in a blockchain can be studied from the perspective of game theory \cite{Kiayias16}.

The main problem of this mining scheme has to do with the huge amount of energy required to support the blockchain network. This issue is raising severe concerns and being a subject of careful \linebreak study \citep{Vranken17}, as can be seen for example in several research works focused on studying the energetic impact of the Bitcoin network \citep{Fairley17}. In the specific case of this cryptocurrency, special hardware (known as ASIC) has been designed and manufactured to accelerate the brute-force process needed for mining.

Some countries where electric supply is cheap (e.g., China and other countries in Asia) are starting to see very large infrastructures exclusively dedicated to Bitcoin or other cryptocurrencies mining. These places are commonly known as ``mining farms'' \citep{Chow17}. Today, the estimation of the amount of electrical power consumed by the Bitcoin network is about 70 TWh per year \citep{BitcoinConsumption}. This amount is similar to the overall consumption of a country like Austria, and, what is worse, all this power is spent in the brute-force process, whose only purpose is to compute the nonces for mining a new block, without any other useful application beyond that scope whatsoever.

For this reason, some authors are proposing a new paradigm in which the mathematical problem required to be solved for mining a new block could have some interest by itself, beyond the usefulness of supporting the blockchain network. In some cases, this approach is called proof-of-useful-work \citep{Li18}, thus recognizing that the computational work done for mining a block must have some interest or useful application.

In this paper, we propose a novel proof-of-useful-work scheme for supporting a blockchain network, whose interest resides on the fact that the mining process is equivalent to the training of artificial intelligence models with many potential applications.

The remainder of this paper is structured as follows: in Section \ref{sec:relwork} the concept of proof-of-useful-work and some attempts to introduce such a concept into a blockchain are described, including some with a scope similar to the one presented in this paper. In Section \ref{sec:theor} some theoretical notions which are of interest for those readers not familiar with the concept of deep learning are provided. In the following four sections we introduce our proposal as follows: in Section \ref{sec:reqs} the requirements that the proof-of-work scheme must follow in order to be implemented within a Blockchain network are described. Section \ref{sec:proc} details the proof-of-useful-work approach proposed in the paper, by which deep learning models are trained and evaluated in order to achieve block mining; Section \ref{sec:map} delves into the process of generating a valid deep learning architecture given a hash; Section \ref{sec:stor} describes an additional proof-of-storage mechanism for being able to support the storage of models and data in a distributed system, and finally in Section \ref{sec:democ} we provide some hints of how artificial intelligence could be democratized in this scheme, by allowing currency owners to propose and vote on problems of interest to the community. Finally, in Section \ref{sec:alt} we propose some possible alternatives regarding the implementation of this conceptual framework, and Section \ref{sec:concl} provides some conclusive remarks about the notions and system proposed in this paper.

%
%
\section{Related Work}
\label{sec:relwork}
Some developers and researchers have already started to work on Blockchain projects which already adhere to the proof-of-useful work scheme, especially in academic settings.

For example, \citet{Zhang17} presented in 2017 the prototype of a system that allows a company to deploy a private blockchain so that they can choose the kind of computation work that must be done in order to append a new block to the blockchain, depending on their needs or preferences.

Similarly, Ball et al. \cite{Ball17,Ball18} have proposed proof-of-useful-work schemes whose hardness is based on different computational problems, such as orthogonal vectors, 3SUM or all-pairs shortest paths problems, claiming that the energy spent on solving them is therefore not wasted.

An example of a proof-of-useful-work scheme that has led to the development of a cryptocurrency is PrimeCoin \citep{King13}, whose mining algorithm consists in searching for Cunningham chains and bi-twins chains of prime numbers. The currency is available in many exchanges under the code XPM and as of December 2018 is valued US\$0.18.

When it comes to the combination of artificial intelligence and blockchain, there are many works (including projects beyond academia) claiming to combine both fields of study in very different ways. A common case consists in trying to apply machine learning for crypto-trading, or extensively trying to predict the price of a cryptocurrency using machine learning \citep{Alessandretti18,McNally18}. Other works have focused on the application of artificial intelligence to blockchain security \citep{Dey18}. From a more philosophical perspective, Swan commented about ``blockchain thinking'', the possibility of formulating thinking as a blockchain process \citep{Swan15}.

Some examples of projects arising from the fusion of artificial intelligence and blockchain are DML (standing for decentralized machine learning) \citep{DML} or SingularityNET \citep{Singularity}. These approaches allow distributed machine learning, with the possibility to deploy smart contracts over a blockchain, and there is not a close integration of both ideas. Another related project is AICoin \citep{AICOIN}, although their approach is not very clear. Finally, in the last quarter of 2018 Fetch.AI was presented \citep{FetchAI}, which mentions both useful proof-of-work and artificial intelligence, although again, the technical details are not clear. In any case, to the best of our knowledge, there are no works using deep learning as a proof-of-work scheme.

Finally, in this work we will present an approach to store models in a distributed fashion that we have deemed to call ``proof-of-storage''. This concept is hardly new, and the first approaches go as far back as 2007 \citep{Ateniese07}, and derivative works have been presented under different names, including \linebreak proof-of-space \citep{Ateniese14,Dziembowski15} and proof-of-storage \citep{Kamara13}.

Proof-of-storage models running over blockchain have been recently presented \citep{Li18}, and have led to actual cryptocurrencies, such as Permacoin \citep{Miller14}, Sia \citep{Vorick14} or Filecoin \citep{Filecoin17}.

%
%
\section{Theoretical Background}
\label{sec:theor}
In this section we will introduce the fundamentals of deep learning, which are required to understand this work's proposal and its context.

Deep learning is a rather generic name used to encompass different artificial intelligence techniques that present some common properties: (1) they require large amounts of data to learn useful models, and (2) they are able to automatically extract relevant features from raw (or almost raw) data.

One of the most widely used deep learning techniques are deep neural networks, which are used for solving a variety of problems both in academy and industry. Some of the common implementations of deep neural networks are convolutional neural networks \citep{LeCun98}, which are often used for computer vision problems, or long-short term memory (LSTM) networks \citep{Hochreiter97}, rather used for audio and natural \linebreak language processing.

In a nutshell, these networks work as follows: a batch of data of a given size is introduced to the network. Then, the first convolutional layer applies a convolution (or most often a cross-correlation) operator over the input, generating an output (called feature maps), which will be used as the input to the following convolutional layer. Optionally, there could be pooling layers in between the convolutional layers, which serve for the purpose of reducing dimensionality by performing subsampling. After the input is consumed by all the convolutional layers, the output feature maps will be flattened into a vector and introduced into a classifier, which can be a classical feed-forward network (such as a multi-layer perceptron) or a recurrent network. Once a classification is obtained, it is compared to the actual value and then the network weights are upgraded using backpropagation.

It is worth mentioning that the previous process describes the most typical type of convolutional network, although there are different implementations with some particular features. For example, fully convolutional networks \citep{Shelhamer17} use additional convolutional and pooling layers for classification, instead of a fully-connected subnetwork; or residual networks \citep{He16} incorporate additional passes of feature maps between non-sequential layers.

Regardless of the specific architecture, a common feature of all deep neural networks is that there exist a remarkable amount of hyperparameters that must be decided upon, to mention a few: the batch size, the number of convolutional layers and of fully-connected layers, the number of convolutional kernels and their size, the pooling setup, the number of neurons in fully-connected or recurrent layers, or the learning setup (optimizer, learning rate, etc.).

The particular setup of these hyperparameters could have an important effect on the neural network performance. Unfortunately, there are no systematic ways to determine the best allocation of hyperparameters, so a common approach is to test different alternatives by trial and error. More recently, improvements in hardware have enabled the automation of this search process, leading to the rise of ``neural architecture search'', where techniques such as reinforcement learning \citep{Bello17} or evolutionary algorithms \citep{Real17,Baldominos18} are applied to optimize the neural network design. In-depth surveys exploring neural architecture search have been published by \citet{Baldominos19} and \citet{Stanley19}.

In this paper, we will not focus on any of these techniques in particular, but rather propose a theoretical framework to carry out neural architecture search in a distributed system, as part of a proof-of-work scheme in a blockchain network supporting a cryptocurrency.

%
%
\section{Formal Requirements}
\label{sec:reqs}
In our proposal, the problem to be solved in order to be able to append a new block to the blockchain consists in the resolution of an artificial intelligence problem. In particular, we propose that these problems require a machine learning model to be trained and evaluated.

The following aspects must be observed for the mining scheme to be able to be introduced into a blockchain network:

\begin{enumerate}[leftmargin=*,labelsep=4.9mm]
    \item The problem must be complex, requiring some computational effort, in order to guarantee that some actual work was performed by miners in order to be able to obtain the reward associated with block mining.
    \item In order to guarantee the integrity of the blockchain, the hash of the previous block must be introduced as a variable of the problem.
    \item The mining scheme must have a competitive component, so that it is the first miner to solve the problem (or conversely, the miner who provides the best solution) the one that mines the block and obtains the reward.
    \item Given a problem solution, it must be easy to verify that the solution is valid and to assess its quality.
    \item Once a miner has found a block and this has been appended to the blockchain, all other potential blocks being under mining by other miners must be discarded. This guarantees that a miner cannot ``save blocks'' to be discovered in the future.
\end{enumerate}

%
%
\section{Proof of Useful Work}
\label{sec:proc}
The purpose of the proposal contained in this section is the description of a novel blockchain network that sustains an alternative coin, namely ``Coin.AI''. Of course, the core of the proposal is the proof-of-useful-work mechanism, which is described in this section so that it complies with the requirements outlined in the previous section.

Regarding Requirement 1, we propose that the problem to solve by miners is the training of an artificial intelligence model, and more particularly a deep learning architecture (probably a convolutional neural network or a variation of it, although the specifics can remain open and are not described in this paper). Training a deep learning model out of data involves learning its parameters (weights) using some kind of gradient descent algorithm. This is an iterative process which is generally regarded as a computationally expensive task, therefore fulfilling such requirement.

In order to append a new block to the blockchain, first the hash value from the last block will be taken. Next, the miner will choose a maximum of $N$ transactions from among the set of pending transactions, which are those which have not been yet included into the blockchain and thus remain unconfirmed. The value of $N$ represents the maximum number of transactions which can be stored in a block. Miners will have freedom to choose whichever transactions they want to include in the block, either randomly or according to some policy of their preference, although it is reasonable to expect that the decision will be based on the fee included in the transaction, as it happens in other blockchain networks such as Bitcoin \linebreak or Ethereum.

Once the transactions have been chosen, miners will generate a number of their choice, called the \textit{nonce}. This number can be assumed to be generated randomly, although a miner could choose the number in any other way. Finally, the hash from the previous block will be appended to the list of chosen transactions and the generated nonce, and the digest of this concatenation will be computed using a cryptographically secure hash function, such as one from the SHA-3 family.

The obtained hash value will then determine the hyperparameters of a deep learning architecture which can then be trained to obtain a valid model. In other words, a function must be devised that maps a hash (which is in fact a number) into a definition of the structural features of a deep learning architecture. This function must be surjective, although it might not be injective, meaning that two different hashes might map into the same configuration. Therefore, bijectivity is not a requirement, although it would be advisable that the target set of candidate configurations is very large. By following this process, Requirement 2 is fulfilled, since the hash of the previous block is introduced as a variable to the problem.

Once the deep learning architecture setup is obtained, the miner can start to apply some backpropagation algorithm in order to train the model for a given problem and data. In order to determine which miner finds the block, we suggest a novel proposal: instead of letting the first who obtains a model to win, the decision would be based on who finds the best model, therefore complying with Requirement 3. To choose a winner, a threshold of minimum performance will be established by the platform. This threshold will be reduced as time elapses while no miner is able to provide a compliant model whose performance satisfies such minimum threshold, in order to allow blocks to be mineable in a reasonable time. During this period, all miners can propose one candidate block, formed by the hash of the previous block, the chosen transactions, the nonce, and the snapshot of a trained deep learning model. 

The choice of training hyperparameters is left to the miner. As a result, miners will be forced to find a compromise solution. A poor training, for example one involving few training epochs, will result in a non-competitive model which will hardly be able to exceed the performance threshold. Conversely, a very exhaustive training could not be able to deliver a prompt solution, allowing another miner to submit a valid solution faster and therefore mining the block.

Additionally, even if all candidate setups must be valid, it could happen that the setup resulting from the obtained hash needed some computing requirements that the miner could not satisfy (for example, a higher amount of memory). In this case, it is likely that the miner will know it before starting the training, since many deep learning frameworks try to reserve memory in advance. In such a case, the miner could decide to change the chosen transactions and/or the nonce and try again, until he has enough computational resources to complete the training.

Once a model is delivered and if this exceeds the quality threshold (a minimum performance enforced by the network for the given problem), then it is submitted to a validation process. During this validation, the platform will assess that the chosen setup matches the candidate block (i.e., that the mapping function has been used correctly) and the model performance will be computed. It is important to notice that while training is a very expensive process in terms of time and computational resources, the operation of computing the performance of a serialized model is much cheaper and faster, requiring a fraction of the effort required during training. This property is consistent with Requirement 4 stated in the \linebreak previous section.

If the model does not exceed the quality threshold then it shall not be broadcasted to the network for assessment to save bandwidth. In the case the miner would still submit the model for validation, it would be discarded and the miner would need to try again with a different architecture.

Once the winning model is determined, the block will be appended to the end of the blockchain. By doing so, the hash of the last block changes, becoming now the hash of the freshly mined block. This implies that all other models must be discarded, since the hash change will necessarily imply a change in the candidate setup, and thus in the model architecture. This would satisfy Requirement 5, the last one. Then, the process can start again to begin the mining process of the following block.

\begin{figure}[H]
\centering
\includegraphics[width=.65\textwidth]{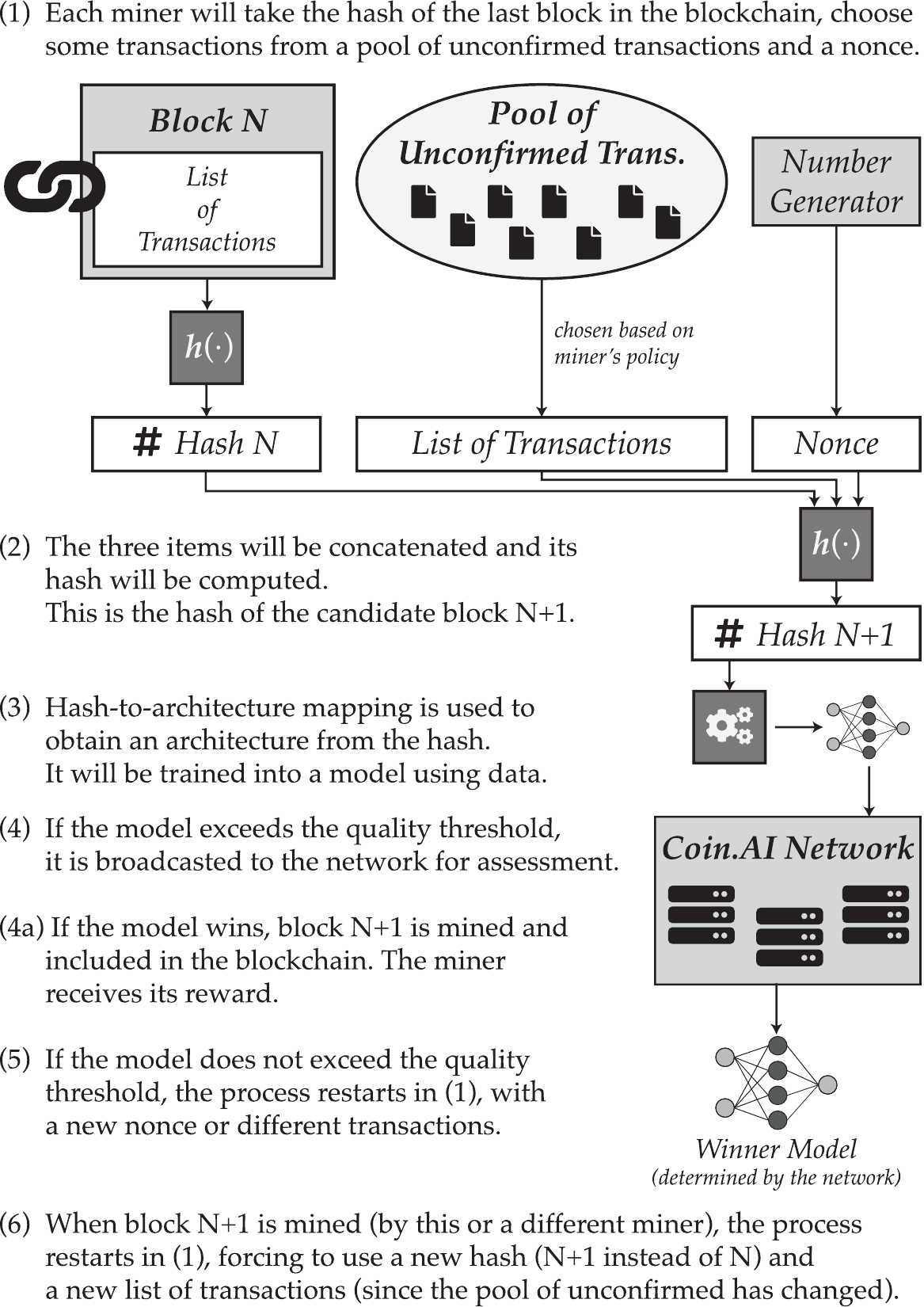}%
\caption{Summary of procedure suggested for the proof-of-useful-work scheme.}
\label{fig:pow}
\end{figure}

The process involved in the proof-of-useful-work scheme is summarized in Figure \ref{fig:pow}.

%
%
\section{Hash-to-Architecture Mapping}
\label{sec:map}
In the previous section we have described the proof-of-useful-work scheme, but the reader has surely noticed how the procedure for generating an architecture setup from a hash is not described. In the current section we describe the mechanism that establishes a surjective function between a hash value and a valid deep learning architecture setup.

In our proposal, let $\mathcal{G} = (V, \Sigma, R, S)$ be a formal context-free grammar defined as a 4-tuple, where $V$ and $\Sigma$ are sets whose union is the whole set of symbols available in the grammar, $V$ being the set of non-terminal symbols and $\Sigma$ being the set of terminal symbols; $R$ is the set of production rules in the grammar and $S \in V$ is the start symbol, which constitutes the root for deriving new strings from the grammar.

$\mathcal{G}$ is the formal grammar that generates a language whose strings are valid setups of a deep learning architecture. In order to generate a string, production rules of the form $\alpha \rightarrow \beta$, where $\alpha \in V$ and $\beta \in (V \cup \Sigma)^*$, are followed in order to eventually generate a valid string $s \in \Sigma^*$. Production rules can be disjunctive, meaning that several rules of the form $\alpha \rightarrow \beta_i$ can co-exist. In addition, rules can be recursive, meaning that for a rule $\alpha \rightarrow \beta$ then it can happen that $\alpha \subset \beta$. Recursive rules are the key for enabling arbitrarily large architectures.

In this theoretical proposal we would like to let the hash-to-architecture mapping remain as generic as possible. By doing so, the grammar to be used could be determined at the time of implementation, and could be modified periodically to adapt to new, more complex deep learning architectures.

However, in this section we present a use case to exemplify the mapping from the hash to an architecture, using a sample grammar. This grammar is not intended to be enforced in an implementation of the blockchain, but is rather presented to illustrate the process.

Let an example of a formal context-free grammar $\mathcal{G}$ be the following, expressed in Backus-Naur Form (BNF). This grammar is inspired in the work by \citet{Baldominos18} and is useful for defining a generic setup for a convolutional neural network:

{\small
\begin{verbatim}
<cnn> ::= <convs><fcs>
<convs> ::= <conv> | <conv><convs>
<fcs> ::= <fc> | <fc><fcs>
<conv> ::= <num_filters><filter_size><act_fn>
<fc> ::= <num_units><act_fn>
<num_filters> ::= <num>
<filt_size> ::= <num>
<num_units> ::= <num>
<act_fn> ::= <sigmoid> | <tanh> | <relu>
<number> ::= <digit> | <digit><number>
<digit> ::= 0 | 1 | ... | 9
\end{verbatim}}

\begin{figure}[H]
\centering
\includegraphics[width=.65\textwidth]{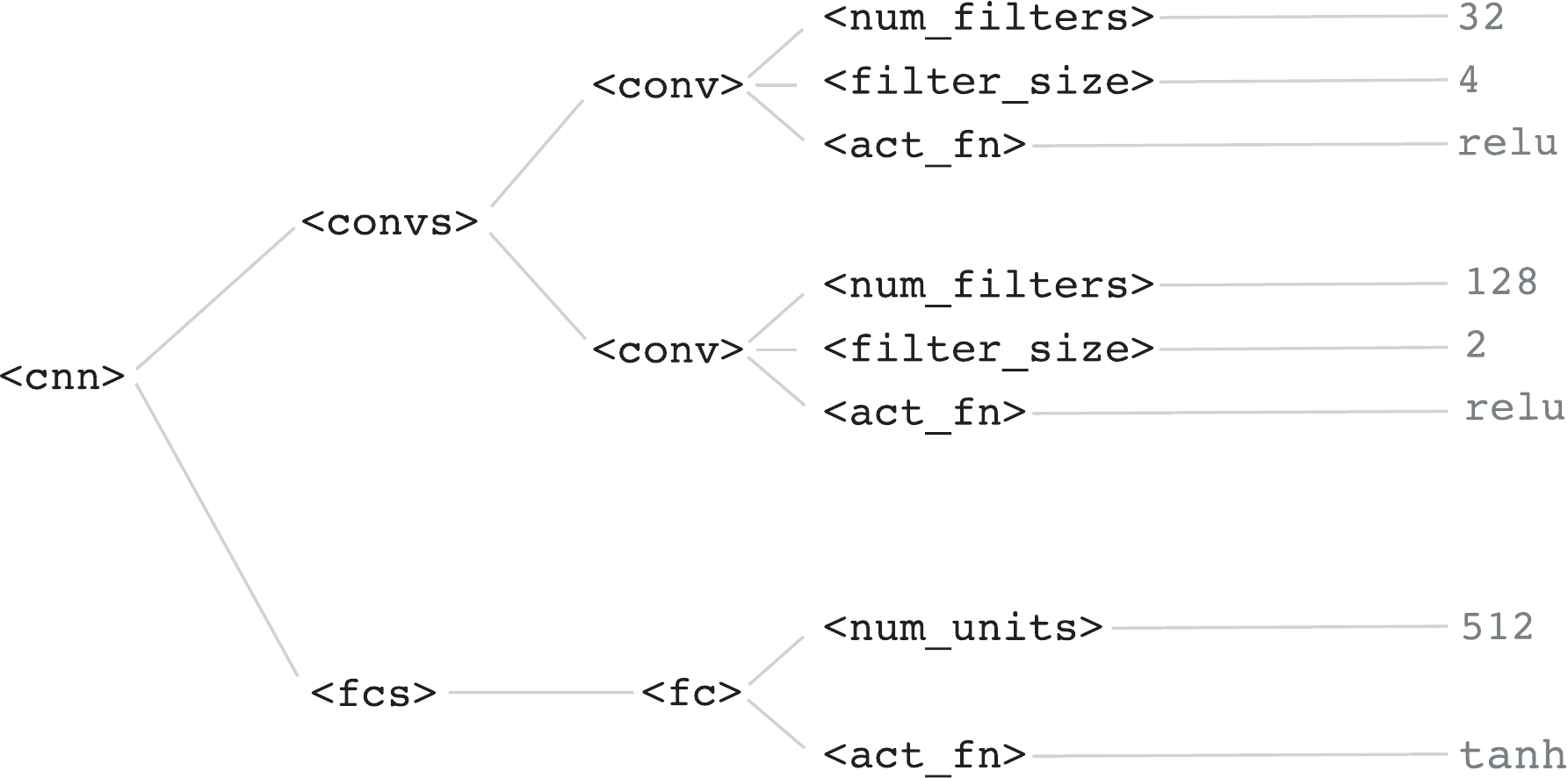}%
\caption{Parsing tree of a CNN architecture using the context-free grammar proposed in the use case.} 
\label{fig:derivtree}
\end{figure}

In this grammar, $S =$ \texttt{<cnn>}, and starting from it a final valid setup can be generated by choosing rules iteratively until a valid string (i.e., composed only of terminal symbols) is obtained. An example of an architecture obtained using this grammar is shown in the parsing tree in Figure \ref{fig:derivtree}, consisting on two convolutional layers and one fully-connected layer. The first convolutional layer would have 32 filters of size 4, and the second would have 128 filters of size 2, both of them with a ReLU activation function. The fully-connected layer would comprise 512 units and an hyperbolic tangent activation function.

Once the grammar is defined, let $H = h(\cdot)$ be the hash value resulting from applying the hash function over the concatenation of the previous block hash, the list of transactions and the nonce. Then $H$ is a number, for example, if the chosen hash function is SHA3-512, then $H$ is a 512-bit number. Then, we can use this number for generating a string given the grammar.

To do so, we will first let $m_0 = H$. Then, given the set of rules of the type $\alpha \rightarrow \beta_i, i \in [0, n]$, we can choose one from among such rules by computing $i = m_t \text{ mod } n$. Once a rule is chosen, a new value of $m$ is computed as $m_{t+1} = \left \lfloor{m_t/n}\right\rfloor$.

\begin{figure}[H]
\centering
\includegraphics[width=.65\textwidth]{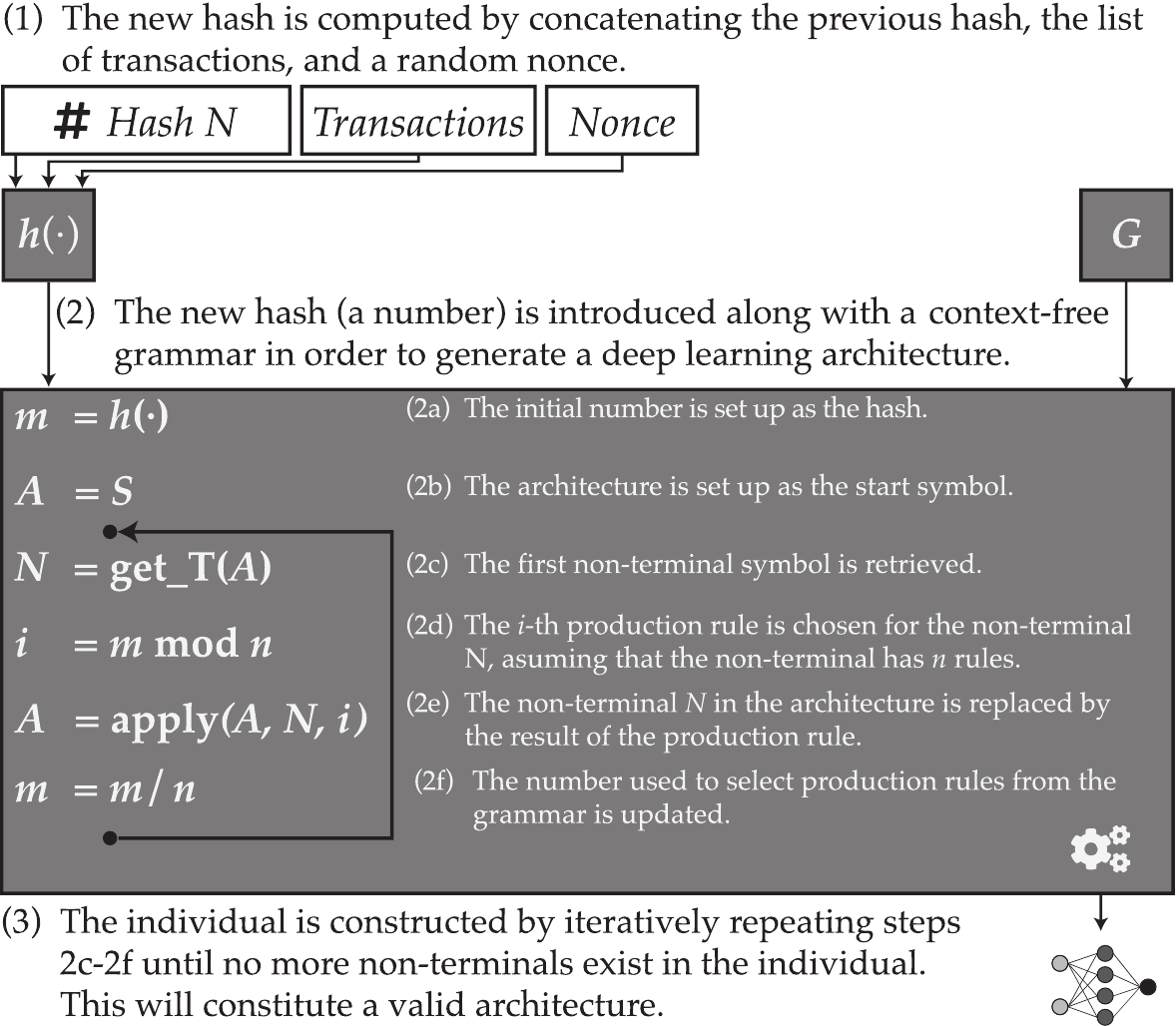}%
\caption{Summary of the process used to obtain a valid deep learning architecture from a blockchain hash by means of a formal context-free grammar.}
\label{fig:h2a}
\end{figure}

This process is repeated iteratively, by choosing one rule at a time and decreasing the value $m_t$ until a final string is obtained. In the case that $m_t < i-1$, then $m_t = H$, meaning that the process continues by restarting the number to the original hash value. A summary of the process is depicted in Figure \ref{fig:h2a}.

%
%
\section{Proof of Storage}
\label{sec:stor}
An additional issue that arises with the previously described proof-of-work scheme is where to store the winning model, i.e., that delivered by the miner who has successfully mined the block. At this point, it is worth recalling that all other non-winner models are discarded.

Currently, three possible alternatives could be considered for carrying out the storage:

\begin{enumerate}[leftmargin=*,labelsep=4.9mm]
    \item The model is stored directly in the blockchain, along with the block to which it belongs.
    \item The model is stored in a centralized server (called \textit{models store}).
    \item The model is stored in a distributed system.
\end{enumerate}

The first option has associated a very important handicap, which is that the serialized model would take, in most cases, a very high ratio of the block storage, quite likely more than 99.9\% of it. For such reason this option would become unfeasible, since network nodes would become unable to synchronize the blockchain because of the lack of storage resources.

The second option could be suitable in a private blockchain setup, when implemented within a company or institution. Conversely, if the blockchain is public (such as Bitcoin or Ethereum), it would be advisable that models be stored in a decentralized fashion, therefore avoiding this option.

Finally, the third option can become more interesting within the scope of a public decentralized blockchain. In particular, in our proposal we suggest that only certain blockchain nodes store the models. To resolve this matter, we propose the system described in the remainder of this section.

Once a block is mined, a hash of the serialized model associated to that block will be computed, and this hash will be stored in the blockchain. Following this process, the block will be submitted to a given number of nodes (named \textit{keepers}) that will store the model.

Keepers are users in the blockchain network that are available for storing deep learning models. In fact, their work is similar to those of miners, but instead of providing processing power to the network they will provide storage capability. Blockchain based storage models are not particularly new, since they have been adopted by some alternative cryptocurrencies such as Siacoin. Additionally, just as it happens with miners, keepers will be rewarded for their work, in the form of some value of the cryptocurrency supported by the blockchain.

The distributed storage system would somehow resemble that available in some distributed file system such as Hadoop: for each deep learning model, $R$ replicas would be made of the model which would be sent to $R$ different keepers, trying to balance the load proportionally to the storage provided by each of these keepers. The parameter $R$ is called the replication factor. If a keeper leaves the blockchain network, then the network must create a new replica and sent it out to a different keeper in order to comply with this factor. This mechanism provides the redundancy needed to avoid loosing information.

\begin{figure}[H]
\centering
\includegraphics[width=.7\textwidth]{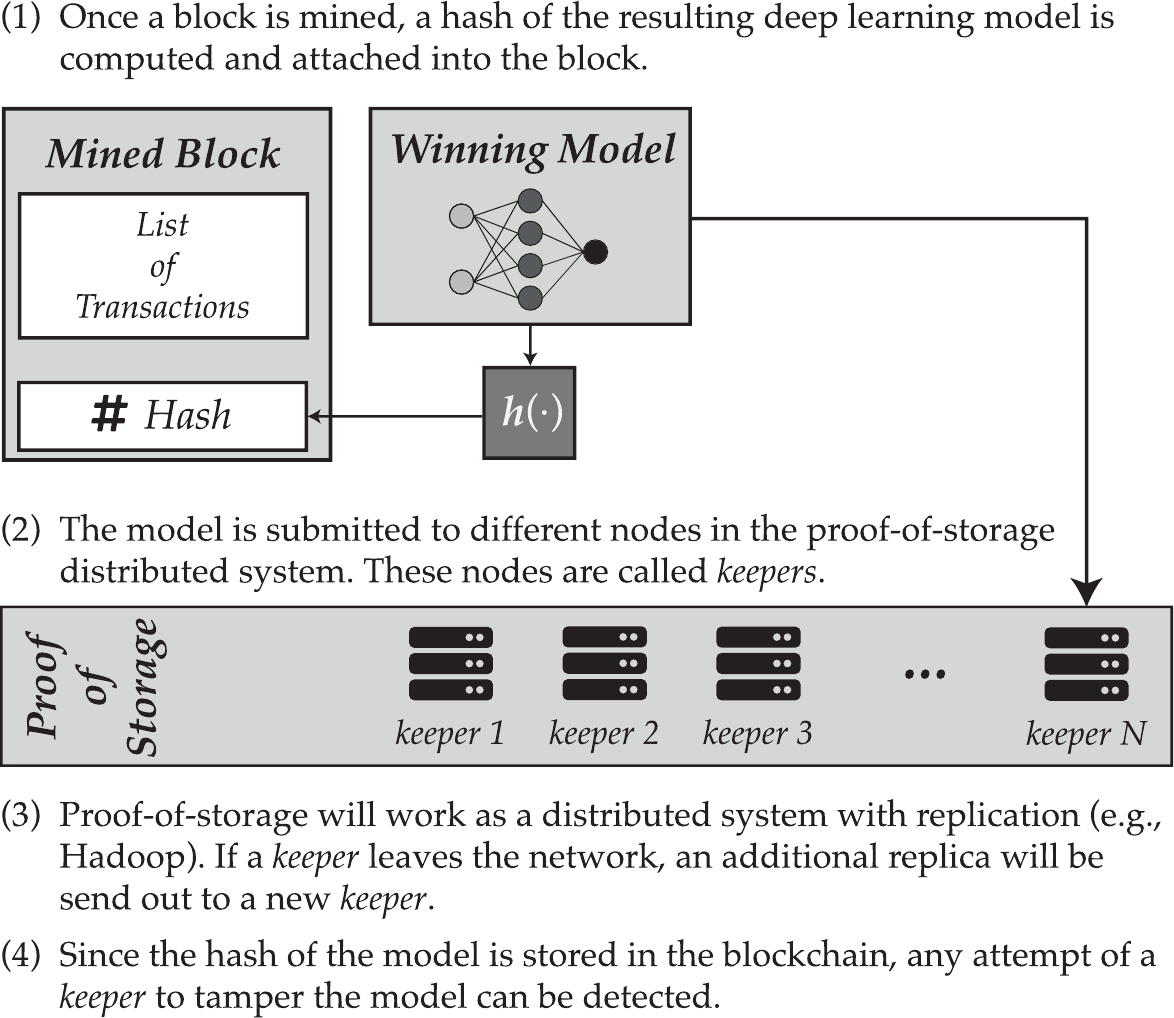}%
\caption{Summary of procedure suggested for the proof-of-storage scheme.}
\label{fig:pos}
\end{figure}

This scheme guarantees a distributed storage of the trained deep learning models. At the same time, since a hash of the model is stored in the blockchain, any attempt of a keeper to tamper a local copy of the model would be detected and could be exposed, considering that replica as invalid and even allowing to penalize the keeper for such a behavior. A summary of how the proof-of-storage scheme is suggested to work is displayed in Figure \ref{fig:pos}.

Finally, although we have so far focused exclusively on the mechanism for storing deep learning models, this system could also be used in order to store the data needed to be able to train such models. In this case, since miners should be able to have fast access to the data in order to train models during the mining process, a higher replication factor could be used. In addition, local copies of the data could be cached by the miners. This is feasible since data will likely take much less space than all the history of models, and also trying to tamper the data is not useful since the quality of models is assessed by other nodes in the network.

%
%
\section{Democratizing Artificial Intelligence}
\label{sec:democ}
At this point, we have proposed technical design and mechanisms that would allow to develop a blockchain network based on a proof-of-useful-work scheme for building deep learning models that solve problems of interest for the community. However, we have deliberately omitted two aspects of the system: the problems to be solved and the grammar which would be used by the hash-to-architecture mapping function. The reason behind this omission is that both aspects must be decided by the community: the cryptocurrency holders.

The first of such aspects involves the artificial intelligence problems that could be solved during the mining problems. In no way would we be able to know in advance which problems are worth solving, since this decision will be taken based on the interests of the community and will likely change over time. The interest of the community could depend on who constitutes such community, but should ideally be aligned with those interests of the scientific community or even of the society at large. In addition, by allowing holders to be able to choose problems of their interest to be solved, we try to provide an added value to the currency, therefore giving it stability and long-term feasibility.

The mechanism for choosing the problem must be carefully devised. To this extent, we propose that any currency holder, even when he owns a very small amount of the cryptocurrency, can propose a new problem to be solved. By referring to a ``problem'' we actually mean a training set and a validation set along with a quality metric to be computed over the latter, which can then be used by miners to learn the models and by other nodes in the network to validate them.

Besides proposing a new problem, users shall be able to support others' proposals by joining or voting on them. When enough interest exists for switching to a new problem, the platform would decide on a new problem randomly, yet proportionally to the total amount of currency owned by the supporters of each problem. By doing so, we provide democratic access to the network resources to anyone that belongs to the network, while providing an incentive to those users who have a larger amount of currency.

An identical approach would be followed in order to determine the grammar that feeds the hash-to-architecture mapping function. It would not be clever trying to devise a grammar at this point, since scientific progress will lead to the appearance of novel architectures that are not yet known. For this reason, we suggest that owners become responsible for taking part in a democratic process that would upgrade the grammar supporting this mapping function in order to comply with novel architectures.

By doing this, we hope that the platform can provide democratic access to artificial intelligence. Certainly, owners of larger amounts of the currency would be rewarded by giving them a higher chance to see their projects selected by the platform. However, in an ideal scenario the currency would be distributed among many owners and none of them would have a substantial amount of the market capitalization. In that case, candidate problems which are of special interest to the community could be highly supported, and therefore would have a higher chance of being chosen.

%
%
\section{Alternatives}
\label{sec:alt}
Regarding the description of this proposal, there are some aspects where we have considered alternative approaches. Some alternatives that would ease the project feasibility in a first stage were described previously. Nonetheless, other alternatives are discussed here; first, with reference to the proof-of-work algorithm, we could consider an adaptive difficulty by regulating the performance threshold. Therefore, it could be interesting to raise the threshold every time a block is mined, in order to guarantee the improvement of model performance. The main issue of this approach is that because deep learning models are complex and follow non-linear stochastic behaviors, it is impossible to quantify how the difficulty is increasing at each point, and it could happen that even a slight increase of the threshold leads to an unsolvable problem. To handle this case, we propose another variation which involves raising the threshold after each block is mined, and then lowering at a slower pace as time goes on, to ensure that eventually the threshold is achievable.

One of the problems that can be found in current cryptocurrencies is that transaction fees can become very high, and since fees are optional, then transactions with low fees might never be chosen by miners. While this might be a desirable feature from an economical perspective in high-demand scenarios \cite{Easley19}, this could effectively lead to raising inequality \cite{Digiconomist}. The main reason is that large fixed fees in a transaction would prevent the use of the currency to purchase inexpensive or daily goods, since fees would effectively consume most of the money involved in the transaction. As an example, in December 2018 the suggested fee in Bitcoin to have the transaction included in a block in a reasonable amount of time was about \linebreak US\$35 \cite{Fees}. Such a large fee would make the currency worthless to the working or middle classes, and make it useful only to purchase high-valued goods (e.g., real estate) or for financial transactions involving large amounts of money (stock, currency exchange, etc.).

In this work, we might wonder whether we could reduce this ``wealth gap'' by increasing the chances of low-fees transactions to be included in a block. A possible way to do so is to remove the nonce, in that case, a miner would be required to change the set of chosen transactions needed to test a new architecture. This could, however, constitute an important issue if there were few transactions, because maybe there are not so many combinations to choose from, therefore reducing the search space. As a fix to this problem, we could devise the range of possible nonces in a manner that it enhances testing of different combination of transactions.

When it comes to determining the neural architecture, we have proposed using generative grammars. We found this approach to be flexible yet easy to understand and implement. Additionally, we have also considered the possibility of using a graph description language to generate computational graphs. In this approach, the search space would be much larger, and would potentially allow for the discovery of novel architectures that have never been tested before. This constitutes an advantage, but also a handicap at the same time, since it could happen that a large fraction of the search space is impractical and unable to \linebreak be trained.

Finally, when discussing ideas for democratizing access to artificial intelligence earlier in the paper, we always took the decisions that led to what we considered a better outcome for democracy. Nevertheless, we could acknowledge different alternatives which could be more interesting in certain socioeconomic scenarios. For example, instead of choosing a proposal randomly based on the number of votes, we could switch to an approach more similar to a proof-of-stake scheme, where proposals supported by stakeholders with larger amounts of money would be chosen with a higher likelihood. A similar yet different approach would consist in starting an auction so currency owners could bid for their preferred problem. It should be noted that these approaches could improve overall interest in the currency, although would enlarge the wealth gap, and could make the currency less attractive to owners with fewer tokens. In addition, the system could prevent the problem of changing unless the previous problem had been active for a certain amount of time. This could enhance the chances of getting better models, and therefore provide an additional incentive to propose or support projects. However, the lock-up period should be decided carefully (or even be part of the problem selection stage) in order to avoid it from being just an \linebreak arbitrary number.

Some or all of these alternatives could be considered upon implementation of this conceptual framework, adapting the inner workings of the platform supporting Coin.AI to the desired behavior.

%
%
\section{Conclusions}
\label{sec:concl}
In this paper, we have proposed a novel theoretical framework for a proof-of-useful-work scheme for mining in a blockchain setting, supporting an alternative cryptocurrency we have deemed ``Coin.AI''. The purpose of our proposed proof-of-work mechanism is to do neural architecture search at a massively distributed scale, with potential implications towards advancing the state of the art in artificial intelligence.

The core of the framework is the proof-of-work procedure itself, which must fulfill a list of requirements to be useful within the scope of the blockchain. In our proposed method, the essentials of the process are similar to those carried out in other cryptocurrencies such as blockchain, yet with a substantial difference: the miner of one block is he who manages to train a deep neural network whose performance at a given problem exceeds a threshold.

\begin{figure}[H]
\centering
\includegraphics[width=0.85\textwidth]{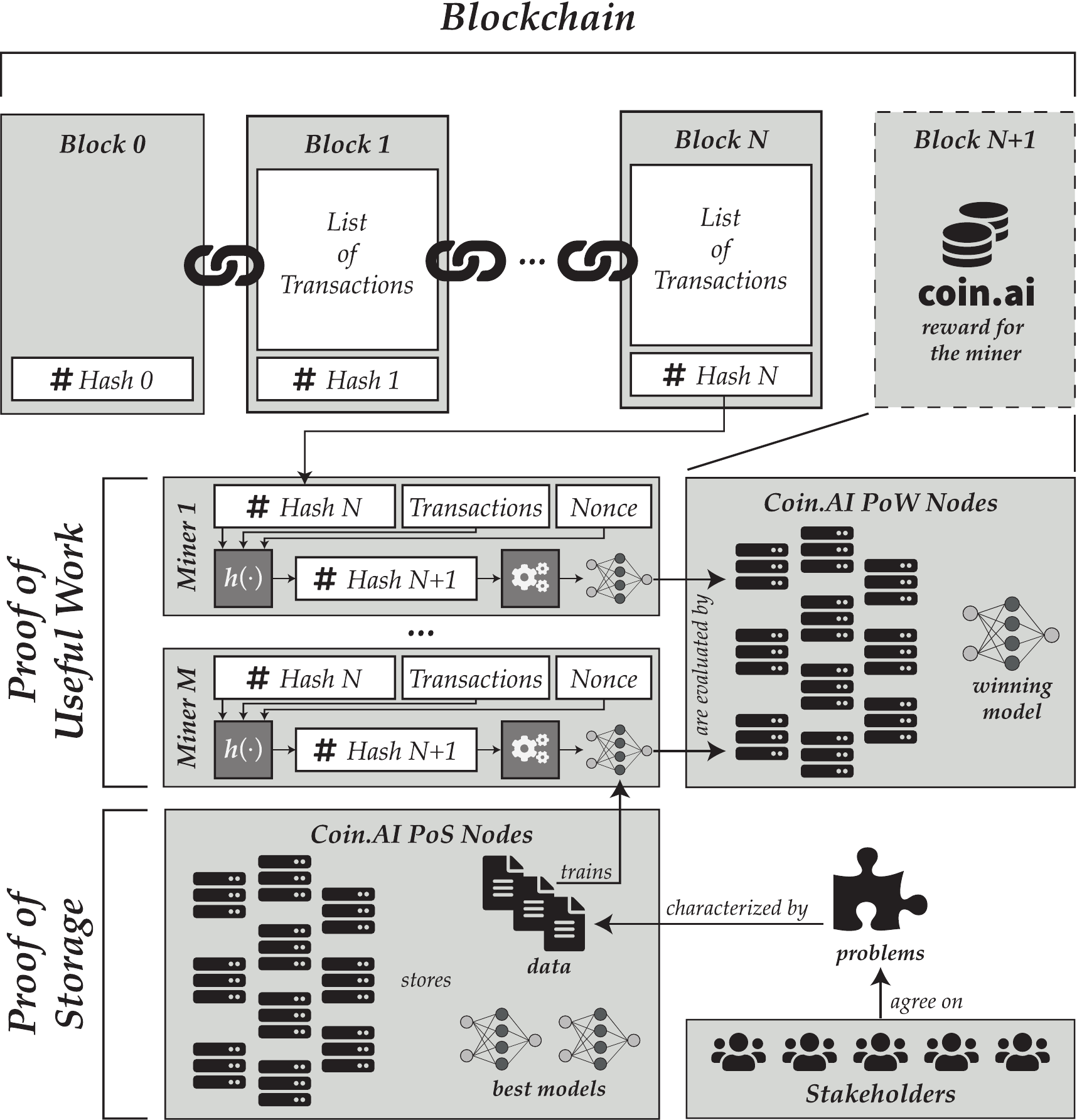}%
\caption{Global overview of the Coin.AI proposal, showing how proof-of-work allows mining a new block and how proof-of-storage is used for storing data and models in a distributed fashion.}
\label{fig:coin}
\end{figure}

The problem of generating a neural architecture during mining is solved by proposing a method that maps the hash of the previous block, the list of transactions and a nonce into such an architecture. This mechanism is fed with a generative grammar, which is used to generate the network architecture from the hash of the aforementioned items. As soon as any element is changed (including the list of transactions or the nonce), the architecture will change as well.
\newpage
Besides the proof-of-work algorithm, we have proposed a proof-of-storage mechanism to securely store the deep learning models in a distributed model. This approach has been applied successfully before, but to the best of our knowledge it is the first time that is combined with a proof-of-useful work scheme. A summary of how the two schemes are combined and a global overview of the Coin.AI proposal are shown in Figure \ref{fig:coin}.

Finally, we have outlined how this alternative coin could be used to democratize the access to artificial intelligence.

\authorcontributions{Conceptualization, A.B. and Y.S.; formal analysis, A.B.; investigation, A.B.; supervision, A.B. and Y.S.; writing--original draft preparation, A.B.; writing--review and editing, A.B. and Y.S.}

\funding{This research received no external funding.}

\conflictsofinterest{The authors declare no conflict of interest.}

\reftitle{References}


\end{document}